\pdfoutput=1

\documentclass[11pt]{article}

\usepackage{ifthen} 
\newboolean{review}
\setboolean{review}{false}  

\newcommand{\iffinal}[2]{%
  \ifthenelse{\boolean{review}}{#2}{#1}%
}

\ifthenelse{\boolean{review}}%
    {\usepackage[review]{acl}}
    {\usepackage[preprint]{acl}}

\usepackage{times}
\usepackage{latexsym}
\usepackage[T1]{fontenc}
\usepackage[utf8]{inputenc}
\usepackage{microtype}
\usepackage{inconsolata}
\usepackage{graphicx}

\usepackage{amsmath}
\usepackage{amsfonts}
\usepackage{xcolor}

\usepackage{hyperref} 
\usepackage{xstring}  
\usepackage{tabularx}
\usepackage{pgf}
\usepackage{subcaption}
\usepackage{graphicx}
\usepackage{adjustbox}

\usepackage{authblk}

\newcommand{\ttw}[1]{\textit{#1}}

\newcommand{\sectionref}[1]{\hyperref[#1]{Section~\ref{#1}: \nameref{#1}}}
\newcommand{\sref}[1]{\hyperref[#1]{§\ref{#1}}}
\newcommand{\appref}[1]{\hyperref[#1]{Appendix~\ref{#1}}}

\usepackage{tcolorbox}       
\usepackage{multicol}        
\usepackage{afterpage}

\usepackage{enumitem}
\setlist{noitemsep}

\usepackage{comment}

\renewcommand*{\Affilfont}{\normalsize\normalfont}

\title{Constructions are Revealed in Word Distributions}
    
\makeatletter
\renewcommand\AB@affilsepx{\hspace{1em} \protect\Affilfont}
\makeatother

\author[1]{Joshua Rozner}
\author[2]{Leonie Weissweiler}
\author[3]{Kyle Mahowald}
\author[1]{Cory Shain}

\affil[1]{Stanford University}
\affil[2]{Uppsala University}
\affil[3]{The University of Texas at Austin\protect\\
\texttt{\{rozner, cashain\}@stanford.edu}\protect\\
\texttt{leonie.weissweiler@lingfil.uu.se} \quad\quad
\texttt{kyle@utexas.edu}
}

\begin{document}
\maketitle
\begin{abstract}
Construction grammar posits that constructions, or form-meaning pairings, are acquired through experience with language (the distributional learning hypothesis).
But how much information about constructions does this distribution actually contain? 
Corpus-based analyses provide some answers, but text alone cannot answer counterfactual questions about what \emph{caused} a particular word to occur.
This requires computable models of the distribution over strings---namely, pretrained language models (PLMs).
Here, we treat a RoBERTa model as a proxy for this distribution and hypothesize that constructions will be revealed within it as patterns of statistical affinity.
We support this hypothesis experimentally: many constructions are robustly distinguished, including (i) hard cases where semantically distinct constructions are superficially similar, as well as (ii) \emph{schematic} constructions, whose ``slots'' can be filled by abstract word classes.
Despite this success, we also provide qualitative evidence that statistical affinity alone may be insufficient to identify all constructions from text.
Thus, statistical affinity is likely an important, but partial, signal available to learners.\footnote{
\ifthenelse{\boolean{review}}%
    {All code and data are provided at ANONYMIZED.}
    {All code and data are provided at \href{https://github.com/jsrozner/cxs_are_revealed}{https://github.com/jsrozner/cxs\_are\_revealed}.}
}

\end{abstract}

\begin{figure}[t] 
    \centering
    \includegraphics[width=\columnwidth]{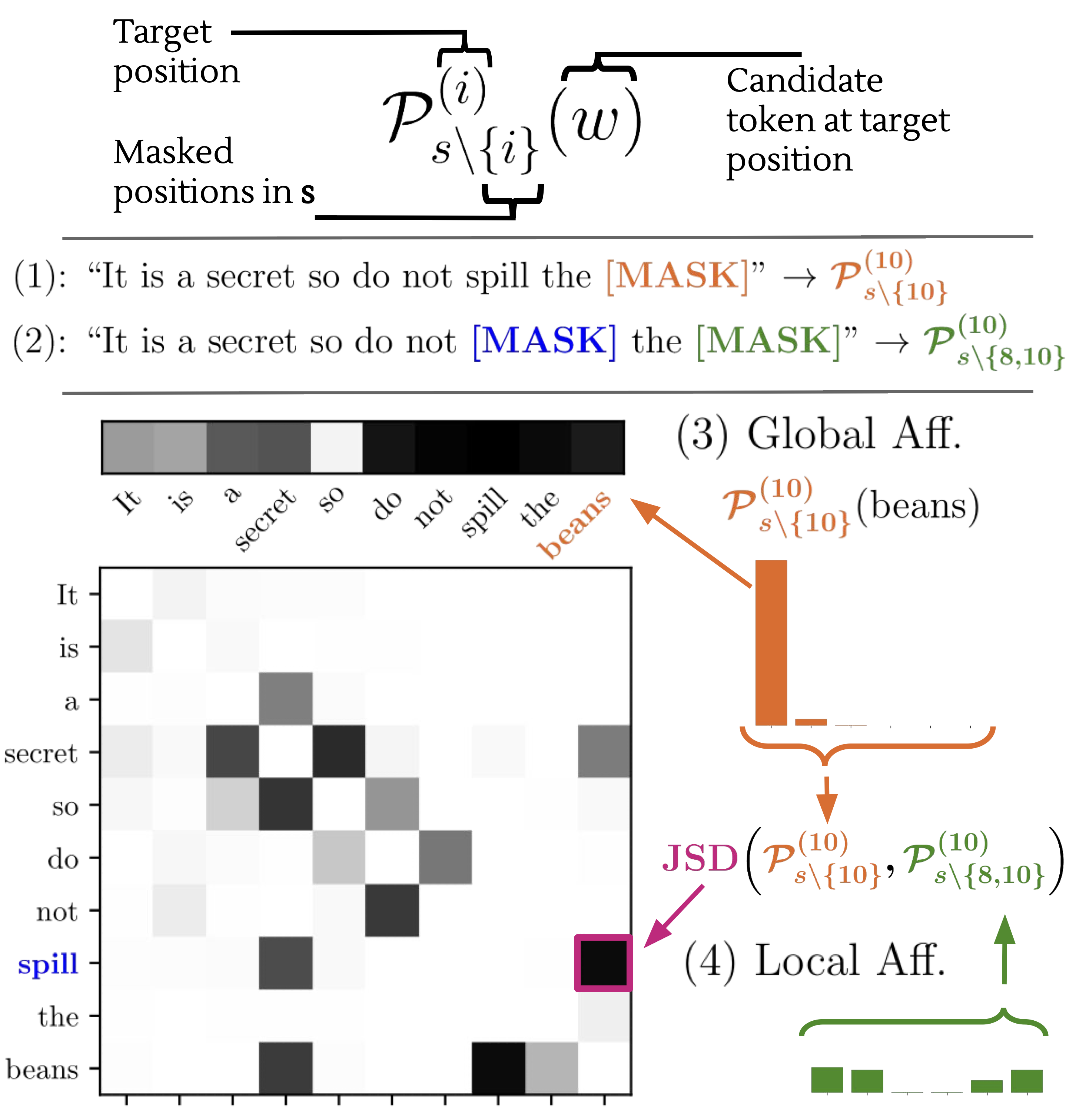} 
    \caption{
In $s = $ ``\ttw{It is a secret so do not spill the \textbf{beans}}'', masking \ttw{beans} (1) gives a 
constrained distribution, 
where $\mathcal{P}^{(10)}_{s \setminus \{10\}}(\ttw{beans})$ is high, so \ttw{beans} has high \emph{global affinity} (3).
By also masking, e.g., \ttw{spill} (2), we get $\mathcal{P}^{(10)}_{s \setminus \{8,10\}}$, compute JSD, and find the words that constrain \ttw{beans} and thus have high \emph{local affinity} (4).
    }
    \label{fig:top}
\end{figure}

\section{Introduction}
\label{intro}

Construction Grammar (CxG, \citealt{goldberg1995constructions, goldberg2003constructions, goldberg2006constructions, fillmore1988mechanisms, croft2001radical}) defines constructions as form-meaning pairings that are acquired through experience with language (distributional learning;
\citealt{goldberg2003constructions,bybee2006usage}).
While the distributional evidence for some constructions is clear (e.g., fixed phrases like \ttw{kick the bucket}), other constructions are less obviously learnable from distributional evidence.
For example, \ttw{I was so happy that I cried} and \ttw{I was so happy that I saw you} are instances of subtly different constructions: they have similar surface form, but opposite causal direction between their component clauses 
(\citealt{zhou-etal-2024-constructions}; see \nameref{related}).

Advocates of CxG have theorized about how children might abstract constructions over time from experience with language \cite{tomasello2005constructing, diessel2004acquisition, diessel2019grammar} and demonstrated the feasibility of distributional learning of constructions in simplified settings \cite{casenhiser2005fast,dunn2017}.
In general, however, we do not have access to the distribution over strings from which children sample.
And though the information contained in this distribution has been characterized using corpus-based methods like collostructional analysis \cite{stefanowitsch2003,stefanowitsch2005covarying,hilpert2014collostructional},
text-only methods do not enable counterfactual questions about what \emph{caused} a particular word to occur in a particular position.
But with dramatic recent advances in statistical modeling of language \cite{zhao2023survey}, we now have pretrained language models (PLMs) that directly instantiate (to a good approximation) the distribution of interest, allowing us to ask how constructions are encoded in statistical relationships between words.

A growing literature explores the use of PLMs as tools for testing usage-based linguistic theories \cite{weissweiler-etal-2023-construction,  goldberg2024, milliere2024LMs, Futrell_Mahowald_2025}.
Two characteristics of this literature motivate the current study.
First, studies to-date have largely been interested in PLMs as \emph{simulations of the learner}, whereas we are interested in PLMs as \emph{simulations of the distribution} from which a learner samples.
These perspectives are not mutually incompatible, but they naturally prioritize different types of analyses.
PLM-as-learner prioritizes questions about the model's behavior (studied using, e.g., prompting, \citealt{zhou-etal-2024-constructions, scivetti2025harish}) or representational geometry (studied using e.g., probing, \citealt{garcia-etal-2021-probing}).
PLM-as-distribution prioritizes questions about how the probabilities over words are influenced by context, irrespective of the model's internal computations (e.g., hidden states).
CxG studies that take this perspective are less common \cite[cf.,][]{veenboer-bloem-2023-using}, and none to our knowledge use causal methods.
Second, current evidence on the learnability of constructions by PLMs is mixed, with some studies reporting success \cite{potts2023characterizing,mahowald-2023-discerning,misra-mahowald-2024-language} and others failure \cite{zhou-etal-2024-constructions,bonial-tayyar-madabushi-2024-construction, scivetti2025harish, weissweiler2024hybrid}.

To address these limitations, we draw inspiration from two areas of research: collostructional analysis \cite{stefanowitsch2003} and intervention methods \citep[see, e.g.,][]{feder-etal-2021-causalm, geiger2022inducing}.
Collostructional analysis measures the statistical affinities that constructions induce between lexical items in a corpus.
Intervention methods systematically alter inputs or hidden states and examine the effects on model behavior.
In this study, we extend perturbed masking \citep{wu-etal-2020-perturbed, hoover-etal-2021-linguistic} to develop \emph{affinity} methods, 
which leverage PLMs as computable models of the language distribution, thereby extending the correlational methods of collostructional analysis to \emph{counterfactual} questions about what causes a particular word to occur.

Our core hypothesis---that constructions will be revealed in the distribution---is partially motivated by idioms, which are loosely defined as semi-fixed multi-word expressions with non-compositional meaning (\citealt{nunberg1994idioms}; \citealt[][p.~248-53]{croft2004cognitive}; \citealt{espinal2019idioms}).
When an idiom is ``activated'' by the surrounding semantics, any compositional or conventional reading is precluded \citep[][p.~169]{hoffmann2022construction}, 
thus \emph{constraining} the slots of the idiom and motivating our notion of \emph{global affinity}; see \nameref{methods}.
Some prior work has investigated how PLMs capture the non-compositional aspects of idioms \cite{zeng-bhat-2021-idiomatic, socolof-etal-2022-characterizing,he-etal-2025-investigating}.
Since many constructions exhibit some degree of non-compositionality \citep[][p.~248--253]{croft2004cognitive}, 
methods that reveal \emph{constraints} in the distribution might recover a variety of constructions (\citealt[][]{croft2004cognitive}; \citealt{taylor2004construction,wulff-oxcg-idioms}).

Using a PLM \citep[RoBERTa;][]{liu2019roberta} as a simulation model, we show that affinity methods---using only the PLM's distribution---recover constructional information across diverse construction types, 
including in previously reported failure cases \citep[cf.][]{zhou-etal-2024-constructions} and across the constructional spectrum from \emph{substantive} (containing slot(s) that are ``fixed'' to a specific word) to \emph{schematic} (containing slot(s) that admit abstract classes of words) constructions.
Nonetheless, we argue from a combination of first principles and qualitative evidence that this distributional approach is likely insufficient to infer the full \emph{constructicon} from data.
In this study, we claim that constructions are revealed in word distributions via affinity methods, and
we organize our contributions as follows:
\begin{itemize}
    \item \textbf{Extension} of prior work (perturbed masking) as \emph{affinity} methods that reveal constructions as patterns of statistical interaction (\sref{methods})
    \item \textbf{Resolution} of previously reported challenges using the methods (\sref{zhou-revisiting})
    \item \textbf{Generalization} of the methods to a wide range of other construction types (\sref{s5-generalize}, \sref{sec:schem})
    \item \textbf{Qualitative analysis} to characterize method behavior and inform the limits of purely distributional approaches (\sref{sec:characterizing})
\end{itemize}

\section{Background}
\label{related}
Prior research on constructions in PLMs has largely used probing or prompting.
Probing has been used to study the representation of constructions in both sentence and contextualized word embeddings (\citealt{weissweiler-etal-2022-better, li-etal-2022-neural}; see also \citealt{weissweiler-etal-2023-construction} for broader survey).
Prompting has been used to elicit acceptability judgements \cite{mahowald-2023-discerning}, semantic understanding \cite{weissweiler2024hybrid}, and constructional similarity judgements \cite{bonial-tayyar-madabushi-2024-construction}.
As these methods test the distributional learning hypothesis only indirectly, they may be susceptible to false negatives and positives.
Successful prompting typically requires models to have logical, metalinguistic, or instruction-following abilities above and beyond basic representation \cite{mccoy-etal-2019-right, mccoy2023embers, basmov2023chatgpt}, and thus prompting might fail to recover constructions that are, in fact, represented in the model's distribution.
Likewise, probing can fail to identify relevant representational distinctions that are mismatched to the design of the probe \cite{adi2017finegrained}, 
or recover distinctions that actually make no causal contribution to the model's behavior \cite{belinkov-2022-probing, hewitt-liang-2019-designing}.

Given these concerns, in this work we emphasize causal relations between input contexts and output distributions, irrespective of how constructions are represented internally by the model, 
since this provides a more direct test of the distributional learning hypothesis.
Some prior work has followed a similar vein:
for example, researchers have used PLMs to score the likelihood of phrases and sentences \cite{hawkins-etal-2020-investigating,misra-mahowald-2024-language} and evaluated semantic effects of constructional context on masked tokens \cite{weissweiler-etal-2022-better,veenboer-bloem-2023-using}.
Our study goes a step further by asking not only \emph{whether} context informs constructional slots, but \emph{how}, by intervening directly on the context itself.

We are motivated to study the distributional encoding of constructions not only by the methodological considerations above, 
buy by prior work on 
challenging constructions that seem difficult to infer from distributions.
For example, \citet{zhou-etal-2024-constructions} report that PLMs fail to distinguish the following three superficially similar but semantically distinct constructions:
\begin{itemize}[label={},leftmargin=2pt]
\item 
    \textbf{Epistemic Adjective Phrase (EAP)} \\
    \hspace*{1em} I was so certain that I saw you.
\item 
    \textbf{Affective Adjective Phrase (AAP)} \\ 
    \hspace*{1em} I was so happy that I was freed.
\item 
    \textbf{Causal Excess Construction (CEC)} \\
    \hspace*{1em} It was so big that it fell over.
\end{itemize}
The general structure is of the form
\begin{equation*}
 [~[\text{NP}] ~[\text{V}] ~\text{so}~ [\text{ADJ}]~]_1 ~\text{that}~[\text{S}]_2
\end{equation*}
where the causal semantics of the three differ: there is no causal relation between $1$ and $2$ in EAP, there is causation from $2 \rightarrow 1$ in AAP, and \emph{vice versa} ($1 \rightarrow 2$) in CEC.
\citeauthor{zhou-etal-2024-constructions} probe and prompt LMs (GPT-3.5/4, \citealt{gpt4}; Llama2, \citealt{llama2})
and argue that LMs fail to reliably distinguish the CEC from the EAP and AAP; see \autoref{zhou-revisiting}.
Relatedly, prior studies of other diverse construction types have suggested that \emph{schematic} constructions with slots for abstract categories or classes, rather than fixed words, may also be especially difficult \cite[e.g.,][]{weissweiler-etal-2022-better}.
In this work, we revisit many of these cases and find that distributions often contain strong signals even for subtle constructional properties.

\section{Methods}
\label{methods}
To test the hypothesis that constructions are revealed as patterns of statistical affinities, we extend perturbed masking \cite{wu-etal-2020-perturbed, hoover-etal-2021-linguistic}, developing
two approaches that compare output distributions under input interventions.
The first is \emph{global affinity}: interaction between a single word and the entire context.
The second is \emph{local affinity}: pairwise interactions between words.

We use the RoBERTa language model \cite{liu2019roberta} for three reasons: (1) it is an open-source, open-weight model with known training data and a \emph{pure} language modeling objective (e.g., no instruction tuning), 
(2) it is a less performant model than those used by e.g., \citet{zhou-etal-2024-constructions}, thus providing a conservative test of the distributional learning hypothesis, and (3) it is bidirectional.
Although bidirectionality is implausible for process models of language comprehension  \cite{frazier1978sausage,elman1990finding,tanenhaus1995integration,altmann2009incrementality,smith2013effect}, 
our goal is not to study processing but the underlying \emph{distribution}, and the constructions we explore here depend on subsequent context.
And though human language learners see much less data than PLMs, they also engage in active learning, taking action in the world and in conversational exchanges \cite{frank2023bridging}; thus a human learner may have some ability to sample more strategically from the overall distribution than a PLM trained passively on text.
For simplicity, we analyze only single-token words, leaving multi-token generalization of these methods to future work.
As RoBERTa has a relatively large vocabulary (50k), we find that this does not pose a substantial limitation for our study.

\subsection{Global Affinity}
Given a string, $s$, of length, $L$, words,
then $s \setminus \mathcal{I}$ is that string with the word indices in $\mathcal{I}$ masked.
Precisely, the \emph{masked string}, $s\setminus\mathcal{I}$, is the string, $s$, with word indices $1 \leq j \leq L$ masked iff $j \in \mathcal{I}$.

Define $\mathcal{P}^{(i)}_{s \setminus \mathcal{I}}$ to be the \emph{probability distribution} given by the model for the $i$th position in the \emph{masked string}, $s\setminus \mathcal{I}$ (note that $i \in \mathcal{I}$).
Then \emph{global affinity} is simply the probability assigned to the original word in the bidirectional context:
\begin{equation*}
\mathcal{P}^{(i)}_{s\setminus\{i\}}(w_i)
\end{equation*}
When a word has high global affinity, the context---potentially involving a construction---strongly informs \cite{shannon1948mathematical} the word's identity.
\autoref{fig:green-day} shows per-word global affinities for the sentence, \ttw{My favorite band is Green Day}.

\subsection{Local Affinity}
Global affinity alone sheds no light on \emph{which parts} of context affect the model's output distribution for a particular word position.
This is limiting for the study of constructions because constructions often involve interactions between multiple slots.
For example, the NPN construction (e.g., \ttw{day by day}, \citealt{jackendoff2008construction}) introduces an interaction between the pair of nouns: the nouns are mutually constrained to be the same.
We quantify such pairwise interactions via \emph{local affinity} between two positions, $i$ and $j$, defined as the distributional difference at position $j$ in a string $s$ as a function of whether the word at position $i$ was masked:

\begin{equation*}
\label{eq:local}
a_{i,j} = \text{JSD}(\mathcal{P}^{(j)}_{s\setminus\{j\}}, \mathcal{P}^{(j)}_{s \setminus \{i, j\}})
\end{equation*}
where JSD represents Jensen-Shannon divergence \cite{lin1991divergence}.
Computing the affinity between each pair of words in an input of length, $n$, words results in an $n \times n$ \emph{affinity matrix}, from which constructions' patterns of affinity can be quantified and visualized (see e.g., Figure~\ref{fig:cec_aff_so_that}).

\section{Revisiting a Challenging Case}
\label{zhou-revisiting}

\newlength{\thwidth}
\setlength{\thwidth}{\dimexpr 0.33\textwidth-\tabcolsep}

\begin{figure*}
\begin{tabular}{@{}p{\thwidth}p{\thwidth}p{\thwidth}@{}}
    \raisebox{-\height}
    {\includegraphics[width=1\linewidth]
{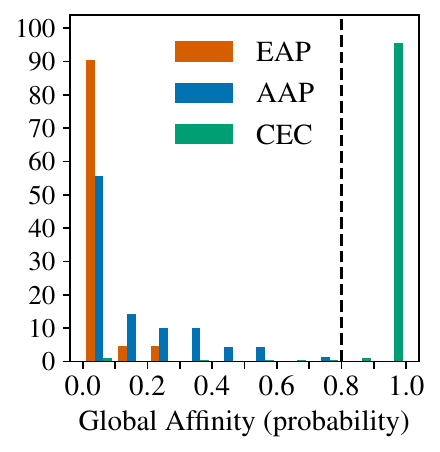}}&
    \raisebox{-\height}
        {\includegraphics[width=1\linewidth]
        {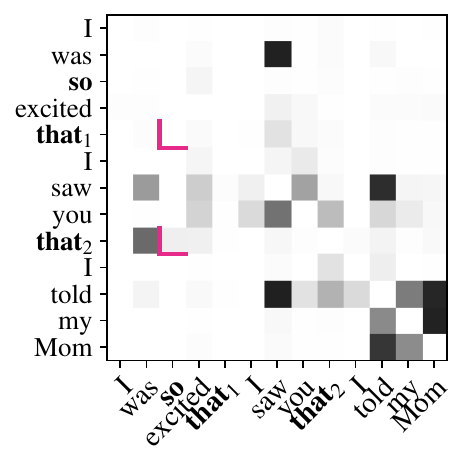}}&
    \raisebox{-\height}
        {\includegraphics[width=1\linewidth]{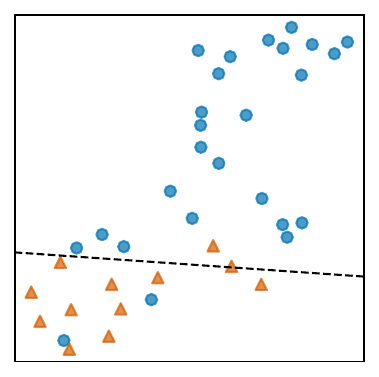}}
    \\
    \caption{
        Percent of examples with \ttw{so} having given global affinity. 
        The CEC is seen to be well-separated from the EAP and AAP.
    }
    \label{fig:cec_hist_so} &
    \caption{
        Local Affinity plot. 
        A \emph{column} shows how much the distribution for that word is affected by other words in the context.
        \ttw{so} is more affected by \ttw{that$_2$} (CEC) than by \ttw{that$_1$}. 
    }
    \label{fig:cec_aff_so_that} &
    \caption{
        UMAP projection (EAP: orange, AAP: blue) using 5 pairwise affinities. 
        Separability with SVM (dashed) suggests interaction patterns differ for EAP and AAP.
    }
    \label{fig:cec_aap_eap_umap}
\end{tabular}
\vspace{-30pt}
\end{figure*}

We first use our methods to address the challenge of distinguishing the CEC from the EAP and AAP, which
\citet{zhou-etal-2024-constructions} test using probing and prompting.
\citeauthor{zhou-etal-2024-constructions} perform classification by probing sentence (GPT-3.5, LLama2) and adjective (Llama2) embeddings.
They use prompting (GPT-3.5/4, Llama2) not for classification, but instead in an experiment that suggests that the models do not understand the causal entailments of the constructions.
In this section, we show that the affinity methods not only robustly distinguish the constructions, but that they also identify mislabeled examples in the original dataset (see ~\sref{cec-unsurprisal}, \sref{cec:eap-aap}).

\subsection{Models distinguish the CEC from the EAP and AAP in their output distributions}
\label{cec-unsurprisal}

Whereas the \ttw{so} in the EAP and AAP can be replaced by other adverbial modifiers (e.g., \ttw{very}), \ttw{so} is required for the CEC to be grammatical \cite{KaySag2012,zhou-etal-2024-constructions}.
If RoBERTa distinguishes the CEC from the EAP and AAP, then \ttw{so} should be constrained (have high global affinity) in the CEC but not in the EAP and AAP.

We calculate the global affinity for \ttw{so} in each sentence and observe that the score distinguishes the CEC:
thresholding global affinity at 0.78 correctly characterizes 272/277 sentences (98.2\%).
In fact, \autoref{fig:cec_hist_so} shows that there is a wide margin, since any threshold between 0.6 and 0.9 achieves similar separation.
\citet{zhou-etal-2024-constructions} separately classify CEC vs. EAP/AAP and report percentage accuracies of 
79.3, 86.5, 68.5 for CEC vs. EAP and 
78.8, 86.5, 68.0 for CEC vs. AAP using
GPT, Llama, and LLama-adj, respectively.
Given the strength of our results, we did not reimplement their procedure; it is clear that our \emph{untrained} approach produces a better result.\footnote{
In fact, \citeauthor{zhou-etal-2024-constructions} upsample the underrepresented EAP and AAP classes.
As our threshold correctly categorizes 100\% of EAP and AAP examples---miscategorizing only 5 CEC examples---
upsampling would \emph{improve} our accuracy.
}

Out of 277 examples, 11 originally appeared to be misclassified using the 0.78 threshold, but upon review, three were mislabeled and three were invalid. 
For example, \ttw{``This was so funny that I had to buy another copy and read it to my better half,''} was originally labeled AAP.
We report results and plots with labels corrected; we provide details of corrected examples in \appref{app:cec-unsurprisal}.
The clear distinction in the model's distribution between the CEC and EAP/AAP---which are superficially \emph{indistinguishable}---provides strong evidence that PLMs do in fact ``retrieve and use meanings associated with patterns involving multiple tokens'' \citep[cf.][]{weissweiler-etal-2023-construction}.

\subsection{Models capture causal relations in the CEC}
\label{cec:so-that}

We have shown that global affinity challenges a previously reported failure of PLM construction representation.
Now we push this finding a step further by using affinity matrices to assess how well PLMs capture the causal semantics of the CEC, as revealed by cases involving multiple clausal complements.
Consider \ttw{I was so excited \textbf{that$_1$} I saw you \textbf{that$_2$} I told my Mom.}
Given the requirement of \ttw{so} to license the CEC,
we hypothesize that if the model captures the causal semantics of the CEC,
then \ttw{so} will have a greater affinity to the causal \ttw{that$_2$} than to the affective one (see \autoref{fig:cec_aff_so_that}).

To test this hypothesis in general, we draw CEC instances from \citet{zhou-etal-2024-constructions} 
and insert additional complementizer (\ttw{that}$\ldots$) phrases to create a small \emph{multi-that} dataset of 31 test sentences (see \ref{app:so-that}).
Across all 31 sentences---even one with five \ttw{that}-phrases---we observe perfect correspondence: \ttw{so} always has the highest affinity with the \ttw{that} at the beginning of the causal excess clause.
This result suggests that the distribution both distinguishes the CEC from similar constructions, and also provides signal for the underlying semantics.

\subsection{Affinity patterns distinguish the EAP and AAP}
\label{cec:eap-aap}

Lastly we consider whether the model distinguishes the EAP and AAP.
\citet{zhou-etal-2024-constructions} showed that the EAP and AAP can be reasonably distinguished, reporting classifier accuracies of 77.1, 71.7, 84.3 for GPT, Llama, and Llama-adj, respectively.
However, given that epistemic and affective adjectives do \emph{themselves} differ, has the probe recovered a constructional distinction or has it just recovered the adjective class?
We cannot answer this question via global affinity: unlike the CEC, which is characterized by the fixed slot constraint for \ttw{so}, no words in the EAP/AAP are fixed, and in some cases when the adjective is masked, both the EAP and AAP are possible (e.g., \ttw{I was so (happy | certain) that I saw you}).
We therefore instead compare them by analyzing patterns in the local affinity matrix.

First, we align examples by identifying the following parts common to all inputs:
<subj$_1$, verb$_1$, so, adj, that, subj$_2$, verb$_2$>.
For example in \ttw{I was so happy that I saw you}, we have <\ttw{I}, \ttw{was}, \ttw{so}, \ttw{happy}, \ttw{that}, \ttw{I}, \ttw{saw}>.
For each example we extract all pairwise affinities between these seven positions.
We test the hypothesis that the constructions have distinct signatures of internal interactions by examining whether the two classes cluster together using UMAP \cite{mcinnes2018umap}, a low dimensional projection.
A UMAP projection of all 49 ($7 \times 7$) dimensions does not separate the two classes;
however, this projection considers all affinities equally, even those which may not be related to the construction (see e.g., \autoref{fig:affinities}).
To address this, we identify potentially salient differences in the interaction patterns and produce the UMAP plot in \autoref{fig:cec_aap_eap_umap} using only the five most substantially different affinities across 26 AAP and 14 EAP examples (admittedly few examples; see \appref{app:cec-umap} for details). 

This method, though imperfect, suggests separability using patterns of interaction between parts of the input.
(See \ref{app:cec-umap} for further discussion of separability.)
\citeauthor{zhou-etal-2024-constructions} were moderately successful in distinguishing the EAP and AAP using probing and unsuccessful with prompting.
Our results suggest that the distribution may be able to distinguish EAP and AAP examples, even \emph{without} access to the adjective's identity.

\section{Generalizing to Other Substantive Constructions}
\label{s5-generalize}

\newlength{\thswidth}
\newlength{\thbwidth}
\setlength{\thswidth}{\dimexpr 0.32\textwidth - \tabcolsep}
\setlength{\thbwidth}{\dimexpr 0.35\textwidth - \tabcolsep}

\begin{figure*}
\begin{tabular}{@{} p{\thswidth} p{\thswidth} p{\thbwidth} @{}}
    \raisebox{-\height}
        {\includegraphics[width=1\linewidth]{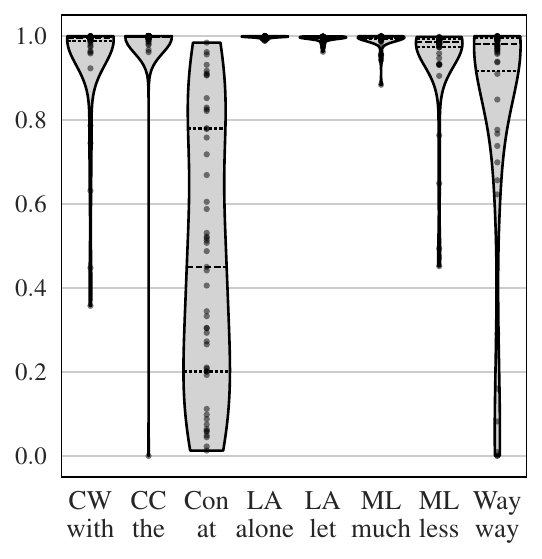}}&
    \raisebox{-\height}
        {\includegraphics[width=1\linewidth]{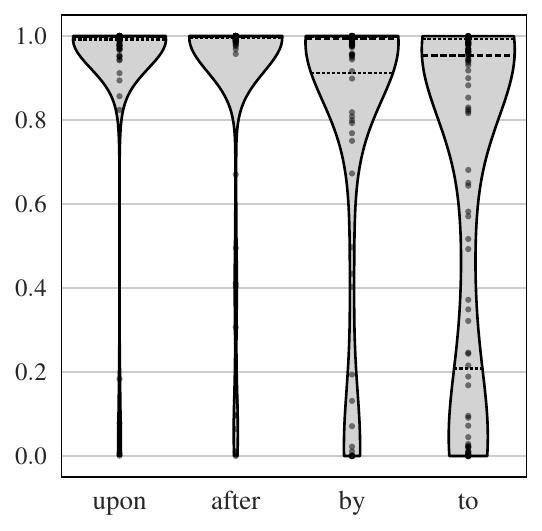}}&
    \raisebox{-\height}
        {\includegraphics[width=1\linewidth]{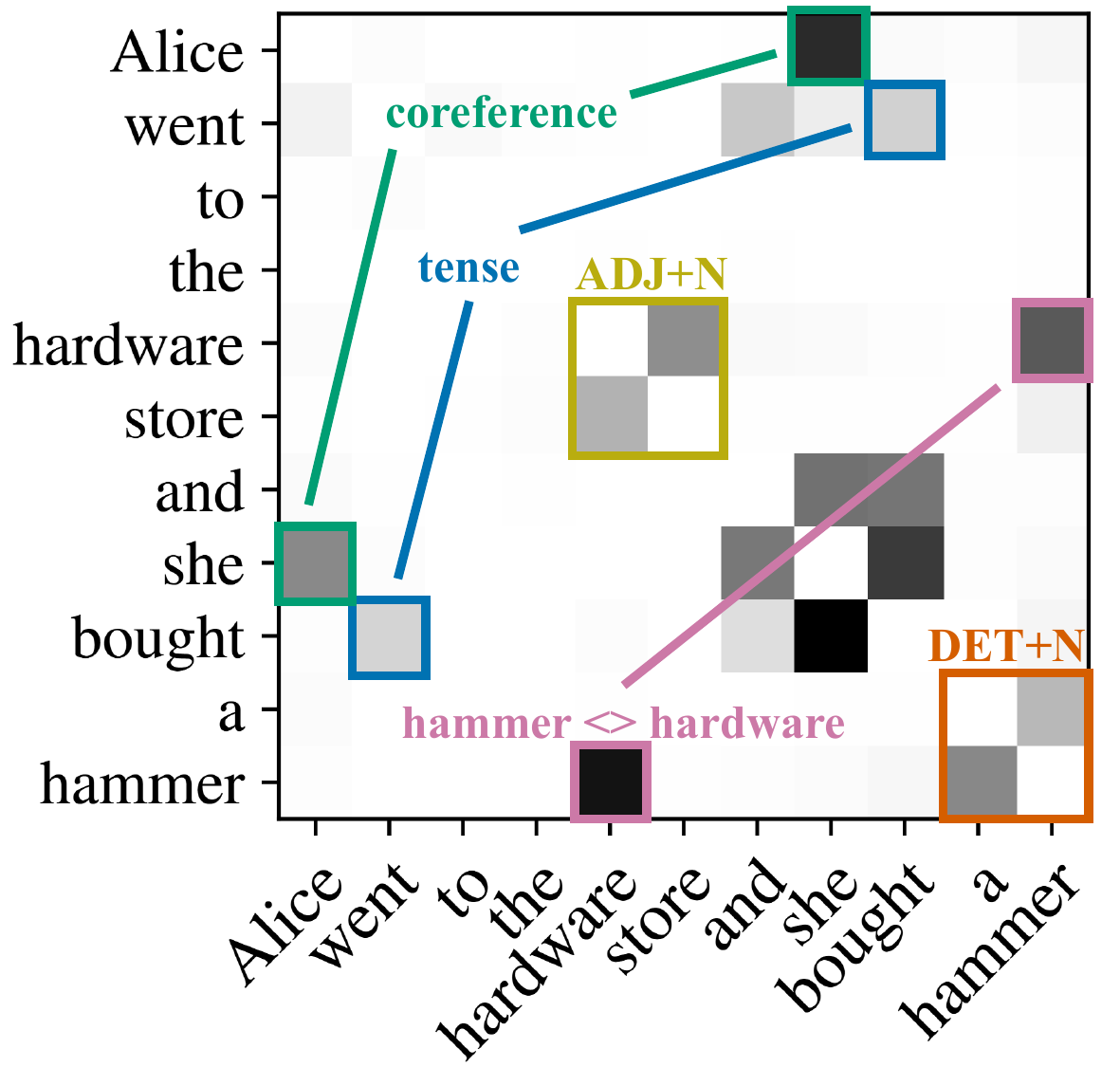}}
    \\
    \caption{
        Global affinity for CoGS. CW: Causative-with; CC: Comp-correlative; Con: Conative; LA: Let-alone; ML: Much-less; Way: Way-manner
    }
    \label{fig:cogs-boxplot} &
    \caption{
        Global affinity for nouns in the NPN construction, grouped by preposition, for sentences with acceptability $\geq4$.
    }
    \label{fig:npn-violin-acceptable} &
    \caption{
        The local affinity matrix encodes diverse types of interactions, including both constructional and non-constructional interactions.
    }
    \label{fig:affinities}
\end{tabular}
\vspace{-30pt}
\end{figure*}

In this and \sref{sec:schem}, we investigate how well our distributional approaches generalize. 
In this section, we study generalization to partially substantive constructions using two datasets: the Construction Grammar Schematicity corpus (CoGS, \citealt{bonial-tayyar-madabushi-2024-construction}) and MAGPIE \cite{haagsma-etal-2020-magpie}, a corpus of potentially idiomatic expressions labeled as either figurative or literal.

\subsection{Global affinity distinguishes fixed slots in numerous constructions}
\label{sec:cogs}

CoGS contains roughly 50 examples for each of 10 construction types.
Six of these constructions are partially substantive and have at least one fixed word for which we calculate global affinity (examples from \citeauthor{bonial-tayyar-madabushi-2024-construction}; fixed words italicized):

\begin{itemize}[label={},leftmargin=2pt]
    \item 
    \textbf{Causative-with:} She loaded the truck \ttw{with} books. 
    \item 
    \textbf{Comparative correlative:} \ttw{The} more \ttw{the} merrier. \\
    (In our analysis the two \ttw{the} words are considered as a single class.)
    \item 
    \textbf{Conative:} He kicked \ttw{at} the ball. 
    \item 
    \textbf{Let-alone:}  None of these arguments is particularly strong, \ttw{let alone} conclusive. 
    \item 
    \textbf{Much-less:}  He has not been put on trial, \ttw{much less} found guilty. 
    \item 
    \textbf{Way-manner:}  We made our \ttw{way} home. 
\end{itemize}

As motivated in the introduction, we expect that constructions will manifest as high global affinities on fixed words.
For example, just as the interactions between words in the CEC constrain \ttw{so},
in the \textbf{let-alone} cxn, we expect interactions to constrain the non-compositional \ttw{let} and \ttw{alone}.

As shown in \autoref{fig:cogs-boxplot}, the fixed words---in all but the conative---have high global affinities.
The \ttw{at} in the conative has low affinity (e.g., \ttw{He kicked \textbf{at} the ball}) because various other non-conative completions are possible: for this example the model produces \ttw{out}, \ttw{at}, \ttw{over}. 
These results show that the distribution simulated by RoBERTa captures the contextual cues associated with various partially substantive constructions, and that global affinity reflects that contextual affinity.

\subsection{Global affinity helps distinguish literal from figurative usages}
\label{sec:magpie}

Next, we ask whether global affinity helps discriminate figurative and literal usages in potentially idiomatic expressions (PIEs). 
We use MAGPIE, a corpus of \mbox{$\sim$50,000} sentences with PIEs that are hand-labeled as either figurative or literal. 
Idioms (e.g., \ttw{kick the bucket}) have long been of interest to CxG \citep[e.g.][]{fillmore1988mechanisms,croft2004cognitive, wulff2008rethinking} and were a key motivation for the approaches in this study (see \nameref{intro}).

We hypothesize that figurative uses will have higher global affinity, by virtue of being entrenched and non-compositional:
consider that one can \ttw{spill the beans} but that in the same context, one would not \ttw{spill the water}, so \ttw{beans} should have high affinity.
Nonetheless, this signal may be confounded: other factors can constrain words (see e.g.,
\sref{sec:characterizing}), and frequent PIEs might have high affinities even when used literally (e.g., ``\ttw{nuts and bolts} in the garage''; see, e.g., \citealt{rambelli-etal-2023-frequent}).

We compute global affinity for each of 114k words that are part of a PIE in 45k sentences (10k literal, 34k figurative; details in \appref{app:magpie}).
Under the hypothesis, we treat affinity as a classification probability for figurative usage
and produce a receiver operating characteristic curve, achieving an area of 0.71 under the curve (plot in \ref{app:magpie}),
which shows that global affinity provides a meaningful signal for classifying figurative vs. literal usages.\footnote{
~``The AUC of a classifier is equivalent to the probability that [it] will rank a randomly chosen positive instance higher than a randomly chosen negative instance,'' \cite{fawcett2006introduction}.
}   

We further compare average figurative and literal scores for each idiom (graphical results in \ref{app:magpie:figlit}).
For example, \ttw{nuts and bolts} behaves as expected, scoring 0.90 fig and 0.84 lit, reflecting entrenchment of the literal usage.
On the other hand \ttw{turn someone's head} gives 0.37 vs. 0.74, but this ``failure'' may stem from the relative non-entrenchment of the figurative use (\ttw{that Yankee sun hasn't turned your head}) versus the more common literal use (\ttw{she turned her head}).
Moreover, qualitative analysis suggests that affinity can be useful to further characterization of PIEs:
for example, the lowest scoring figurative usages are generally less familiar to the authors, 
and affinity helped identify some mislabeled or questionable examples.
Future research combining affinity methods with other approaches might produce further quantitative insights for PIEs.

\section{Generalizing to Schematic Constructions}
\label{sec:schem}

Our analyses thus far have mostly focused on how constructions manifest in affinities for particular \emph{fixed} words in specific slots, but
constructions can also be \emph{schematic} with abstract slots \cite{croft2004cognitive, goldberg2003constructions}.
Here we generalize our approach to study how well the distribution captures two such constructions, the noun-preposition-noun construction (NPN; \citealt{jackendoff2008construction}) and the comparative correlative (CC; \citealt{fillmore1986varieties}).

\subsection{Models generalize the NPN's covarying noun-noun slots}
\label{sec:npn}

The NPN construction (e.g., \ttw{day after day}) is a schematic construction (see e.g., \citealt{sommerer2021absentnpn} for a recent study using collostructional analysis).
Since the construction is entirely schematic (no slots are constrained to be fixed words), a distributional learner can acquire it only by generalizing to abstract classes (i.e., noun and preposition).

To study whether the distribution reflects \emph{generalization} of the NPN to arbitrary nouns,
we randomly sample 100 singly-tokenized, singular nouns from RoBERTa's vocabulary and prompt GPT-4 to produce NPN sentences with each of the prepositions \ttw{by}, \ttw{after}, \ttw{upon}, and \ttw{to},\footnote{
There are other varieties of NPN than the symmetric N+P+N and other prepositions can be used \cite{jackendoff2008construction, sommerer2021absentnpn}; this analysis is intended only to show generalization, not to be an exhaustive study.
}
giving 400 sentences.
We compute global affinities for the nouns in the NPN and obtain acceptability ratings (scale \mbox{1-5}) from the last author, who is blind to the affinity scores.
These ratings are used only to segment our analysis by acceptability.
In \autoref{fig:npn-violin-acceptable} we plot affinity scores for nouns in sentences with acceptability $\geq$ 4  
(upon: 65, after: 73, by: 52, to: 54; total 244).
We include the same plot but without acceptability filtering in the Appendix, \autoref{fig:npn-violin-all}; 
lower acceptability sentences have lower affinity, reflecting that affinity is sensitive to linguistic acceptability.

In \autoref{fig:npn-violin-acceptable}, affinity scores for \ttw{upon} and \ttw{after} suggest that the distribution captures the form of the NPN:
the model generally expects the two nouns to be the same.
Moreover, the lower affinity for NPNs with \ttw{by} and \ttw{to} (and also the relative counts of acceptable generations) accords with prior characterization of the NPN:
\ttw{after} and \ttw{upon} are more flexible in NPN use than other prepositions, and \ttw{to} is only semi-productive \cite{jackendoff2008construction}.

As our objective is to test generalization to unseen NPNs, we generate a separate challenge dataset of $\sim100$ nouns:
Using the infinigram API \cite{Liu2024InfiniGram}, we sample nouns that are not used in an NPN (with \emph{any} of the four prepositions) in the Pile-train dataset (\citealt{gao2020pile}; see \ref{app:npn:infinigram} for details).
Though the affinity distribution for entirely unattested NPNs is more skewed toward lower affinities, we still see clear evidence of generalization (\autoref{fig:npns_challenge_acceptable}).

\subsection{Models generalize the comparative correlative's category constraint}
\label{sec:cc}

We test whether the distribution encodes the semantic category of the comparative adjective/adverb in the CC \citep[see][for a recent study]{weissweiler-etal-2022-better}, which
is of the form, e.g., \ttw{The \textbf{better} your distribution, the \textbf{more} constructions it will encode}.
Whereas in \sref{sec:cogs}, we showed that \ttw{the} in the CC has high global affinity, here we test whether the distribution also encodes an abstract slot constraint for the comparative adjective/adverb.
Using the 54 CC examples from the CoGS dataset, 
we mask each comparative adjective/adverb, obtain the set of highest probability outputs at the masked position that sum to 98\% probability mass, and calculate a \emph{comparative score}: the percentage of this set that is a comparative adjective/adverb (see \ref{app:cc}).

Out of 99 comparative adjectives/adverbs, 95 score 100\%; another 3 score $>99\%$, and one (\ttw{The higher up the \textbf{nicer}!}) scores 86\%, with non-comparatives \ttw{ladder} and \ttw{mountain} in the top 10 fills, perhaps because it is uncommon for a comparative to come at the end of the sentence.
This result shows that the distribution captures the abstract syntactic (comparative adjective/adverb) constraint of the CC nearly perfectly, with 98/99 examples having scores $\geq99\%$.

\section{The Limits of Distributional Analysis}
\label{sec:characterizing}

\begin{figure*}
\begin{tabular}{@{} p{\thwidth} p{\thwidth} p{\thwidth} @{}}
    \raisebox{-\height}
        {\includegraphics[width=1\linewidth]{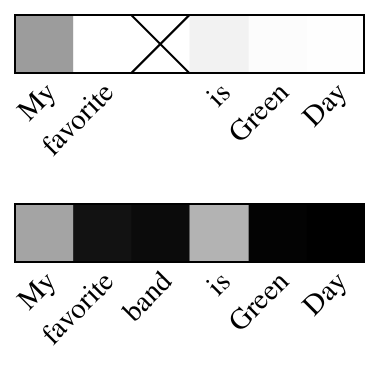}}&
    \raisebox{-\height}
        {\includegraphics[width=1\linewidth]{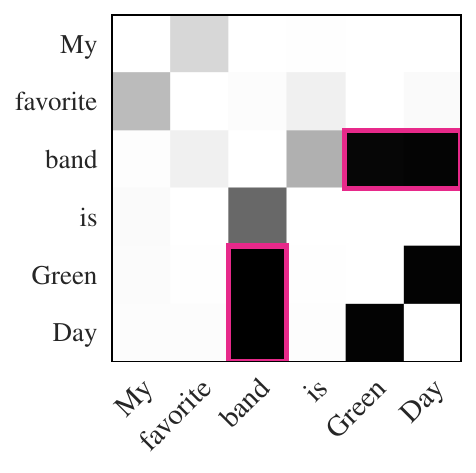}}&
    \raisebox{-\height}
        {\includegraphics[width=1\linewidth]{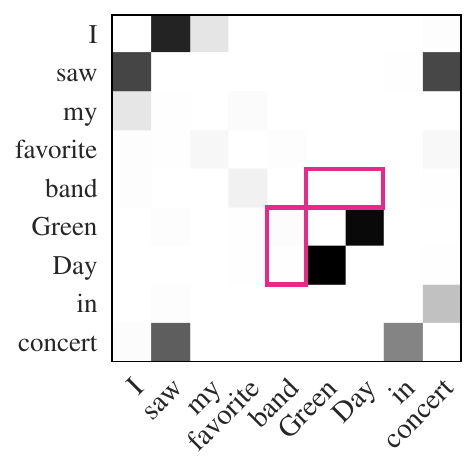}}
    \\
    \caption{
       \ttw{Green Day} (a cxn) is present in top/bottom panels but without/with \ttw{band}, it has low/high global affinity (white = 0, black = 1).
   }
    \label{fig:green-day} &
    \caption{
        The local affinity matrix reflects interactions (pink) between between
        \ttw{band}, \ttw{Green}, and \ttw{Day}, as expected.
    }
    \label{fig:cx_green_band_dist}&
    \caption{
        In contrast with \autoref{fig:cx_green_band_dist}, with additional context (\ttw{\ldots in concert}), affinities between \ttw{band} and \ttw{Green Day} seem to disappear (pink).
    }
    \label{fig:cx_green_day_low_aff}
\end{tabular}
\vspace{-30pt}
\end{figure*}

How far can we push these distributional approaches for identifying constructions in text?
Could they allow us to identify a model's complete constructicon bottom-up \citep[see, e.g.,][]{dunn2017, dunn-2019-frequency,dunn2024computational, feng2022word, lyngfelt2018constructicography, xu-etal-2024-coelm}? 
We argue the answer is likely ``no'': although affinity is a highly useful diagnostic of models' knowledge of constructions, here we show that it cannot be directly 
equated with constructionhood.

First, consider the sentence \ttw{Alice went to the hardware store and bought a hammer}.
The affinity matrix (\autoref{fig:affinities}) reveals  various interactions of interest.
Some of these are likely constructionally mediated, including tense agreement in coordinated verb phrases (\ttw{went}, \ttw{bought}), as well as subject-verb (\ttw{she}, \ttw{bought}), head-modifier (\ttw{hardware}, \ttw{store}), and determiner-noun (\ttw{a}, \ttw{hammer}) dependencies.
But other interactions are less obviously constructional, such as coreference (\ttw{Alice}, \ttw{she}) and semantic relatedness (\ttw{hardware}, \ttw{hammer}).
So although the affinity matrix does reflect constructional relations, it also shows that other contextual interactions can produce high local affinities.
Thus affinity is an insufficient
criterion for constructionhood.\footnote{
Insufficiency holds for \emph{global} affinity as well: \autoref{fig:gaf_alice_hammer} shows that \ttw{hardware} has high global affinity (0.92).
} 

Second, consider the sentence \ttw{My favorite band is Green Day}, which includes the non-compositional collocation \ttw{Green Day} (a well-known band name).
\autoref{fig:green-day} shows that  \ttw{Green Day} has low global affinity until an appropriate contextual trigger is given, i.e., \ttw{band}.
This shows that even substantive constructions may exhibit low global affinities when the surrounding context is insufficient to trigger them.\footnote{
See also \autoref{fig:kick-the-bucket} in the Appendix, for an additional example with the idiom  \ttw{kicked the bucket}. 
}

Additionally,  \autoref{fig:cx_green_band_dist} shows the local affinity matrix for the \ttw{Green Day} example. 
As expected, it reflects affinity between \ttw{band}, \ttw{Green}, and \ttw{Day}.
However, in a different context (\ttw{I saw my favorite band, Green Day, in concert}), the interactions between \ttw{Green Day} and \ttw{band} vanish (\autoref{fig:cx_green_day_low_aff}).
This appears to be due to the co-presence of an additional semantic cue (\ttw{concert}): since \ttw{band} and \ttw{concert} are both individually sufficient to cue the construction \ttw{Green Day}, neither individually exhibits strong pairwise affinity with \ttw{Green Day} (i.e., neither when masked substantially changes the model's output distribution for \ttw{Green} or \ttw{Day}).

Taken together, the \ttw{hardware store} and \ttw{Green Day} examples 
show that affinity is neither sufficient nor necessary for a construction to be present.
Insufficiency arises out of the tension between \emph{contextual} and \emph{constructional} interactions. 
Non-necessity arises from a methodological challenge: eliciting global affinity requires masking, but under mask, the context may not trigger a construction.
These results thus warrant caution in mapping between observables (affinities) and hypothetical constructs (like constructions).
They also suggest avenues for future work, which we explore below.

\section{Discussion}
\label{sec:disc}

We used input interventions to investigate whether constructions result in patterns of statistical \emph{affinities} and thereby manifest in PLMs' output distributions.
Our methods showed that RoBERTa's distribution distinguishes semantically distinct but formally similar constructions that were previously reported as failures, and our approach even identified mislabeled and unclear examples.
We generalized our results to six partially-\emph{substantive} constructions, potentially idiomatic expressions, and two \emph{schematic} constructions that have abstract constraints.
These results support the distributional learning hypothesis: the distribution over strings, as simulated by PLMs, contains rich information about the constructicon.
Nonetheless, we also showed that the distributional measures we developed are, in general, neither necessary nor sufficient to induce constructions.
Instead, statistical affinity is likely one of a broader set of cues, both for linguistic analysis and for language learning.

Compared to prior work, we presented a purely distributional approach to the study of constructions in PLMs:
RoBERTa is a statistical representation of the corpus, obtained via an unsupervised masked-language-modeling objective \cite{devlin-etal-2019-bert}. 
That representation, which encodes a computational model of language, was interrogated via perturbations on strings.
Whereas prior work has used NLI--style entailment queries and bespoke probes trained to identify particular constructions, we instead---and more simply---directly examined the model's output distribution over tokens.

With respect to the open problem of construction induction, our methods may prove useful:
Global affinity can identify \emph{what} is constrained (potentially by a construction), and local affinity can identify \emph{why} it is constrained.
Given that induction over schematic constructions requires assigning semantic categories, our results on the NPN and comparative correlative suggest that the distribution (and thus contextual representations) may encode categories of interest \citep[see also][]{chronis-etal-2023-method}. 

Future work to distinguish \emph{constructional} from \emph{contextual} interactions could be part of an effort to understand constructions information-theoretically \citep[cf.][]{futrell2019syntactic}.
Given recent questions about the falsifiability of CxG \citep[see, e.g.,][]{cappelle2024can}, an information-theoretic approach might provide a quantitative constructional criterion.
Though we studied only \emph{affinity} here, further research might investigate how statistical affinity relates to existing dimensions of constructional analysis like degree of idiomaticity \cite{wulff2008rethinking}, frequency, and entrenchment \citep[][p.~239]{stefanowitsch2003}.

\section{Conclusion}
The distributional learning hypothesis is a fundamental assumption of construction grammar. 
We have shown experimentally that the distribution over strings, as approximated by a PLM, contains rich information about constructions' syntactic and semantic properties.
Across a wide range of construction types, including previously reported hard cases involving semantically distinct but superficially similar constructions, we find that constructional information is reliably reflected in the causal interactions between words and their surrounding context.
This finding both complements existing approaches in linguistics that attempt to characterize constructions using passive text, and supports the hypothesis that distributional information is a major source of signal available to language learners.
Our work offers a methodology that may contribute to the growing field of research on constructions in PLMs, may inform construction induction, and suggests the possibility of a quantitative, information-theoretic approach to modeling constructions.
For linguists our methods offer a new kind of PLM-based approach to corpus study---one which extends existing methods like collostructional analysis to direct counterfactual inquiry.

\section*{Limitations}
Our analysis is limited to a single (bidirectional) masked language model, RoBERTa, and could be rerun on other models of different sizes. 
In choosing masked language models, a straightforward analysis of the sort that we have performed is limited to words that are single tokens;
multi-token generalization of these methods is left to future work.
Our approach for \sref{cec:eap-aap} removed numerous examples, leaving relatively few for analysis; nonetheless the approach still recovered mislabeled examples.
We studied only English constructions, and future work should look to apply these methods to other languages.

CxG is a broad theory  \cite{goldberg2003constructions,goldberg2024,hoffmann2022construction}.
We did not consider the question of precisely defining what a construction is nor did we study any \emph{particular} constructionist approach \citep[see, e.g.,][]{oxcg-2-goldberg-constructionist-approaches}. 
Unlike with other theories of syntax, there is no complete inventory of constructions available, so our study necessarily focused on specific ones that had already been discussed in previous literature.

CxG is furthermore one of many extant theories of natural language syntax \cite[][\emph{inter alia}]{chomsky1995minimalist,pollard1994head,steedman2001syntactic,bresnan2015lexical}.
Although our study targets a key premise of CxG (usage-based learning), we do not claim that CxG is the only appropriate analysis of the phenomena we study, nor are we arguing for CxG over alternative approaches.
Our results take CxG as a starting point and thus do not allow us to weigh in on these important theoretical questions.

\iffinal{
\section*{Acknowledgements}
Joshua Rozner was supported by the Institute for Computational and Mathematical Engineering at Stanford University.
Leonie Weissweiler was supported by a postdoctoral fellowship from the German Research Foundation (DFG, \texttt{WE 7627/1-1}). 
Kyle Mahowald acknowledges funding from NSF CAREER grant 2339729.
We thank Shijia Zhou for providing precise numeric results for prior work on the EAP, AAP, and CEC.
}
{}

\bibliography{latex/anthology-1, latex/anthology-2, latex/josh, latex/literatur_25_03, latex/kyle}

\newpage

\appendix
\label{sec:appendix}

\section{Supplement: Methods}

In this paper we restricted consideration to singly-tokenized words. 
Multi-tokenized words cannot be studied with a single mask (the remaining token would substantially shape the outcome), and
it is not trivial to turn a multivariate distribution over multiple masks into a univariate distribution over words.
In practice, most words we encountered in this study were singly-tokenized and this was not a substantial limitation.

Calculating a local affinity matrix is more expensive than calculating global affinities: $n + n^2$ forward passes versus $n$.
Note that the affinity matrix is asymmetric---positions (i,j) and (j,i) involve three different distributions and are computed as JSD(A,B) vs. JSD(C,B)---and that the diagonal of the matrix, $a_{i,i}$ is $0$.

We use RoBERTa large.
Most experiments are run locally on an M3 Macbook Pro or on a single Nvidia RTX A6000 GPU on a cluster.
Total time to run all experiment code is less than an hour.

\section[Supplement for Section~\ref{zhou-revisiting}]{Supplement for \sectionref{zhou-revisiting}}
\subsection{Data and preprocessing}
\label{app:cec-data}
We preprocess the dataset examples to identify the indices of \ttw{so}, \ttw{that}, and the adjective. 
The original dataset has 323 examples, and our pre-processor and method pipeline fails on 46 examples, leaving 277 for our analysis.
Some of the failures are the result of how our processing interacts with punctuation, others are issues in the dataset.
Particular failures can be obtained by re-running our processing code provided in the repository. 
The final dataset on which we run our analyses has 24 EAP, 70 AAP, and 183 CEC sentences, for a total of 277 sentences.

\subsection[Supplement for Section~\ref{cec-unsurpisal}]{Supplement for \sectionref{cec-unsurprisal}}
\label{app:cec-unsurprisal}
\begin{figure}[t] 
    \centering
    \includegraphics[width=\columnwidth]{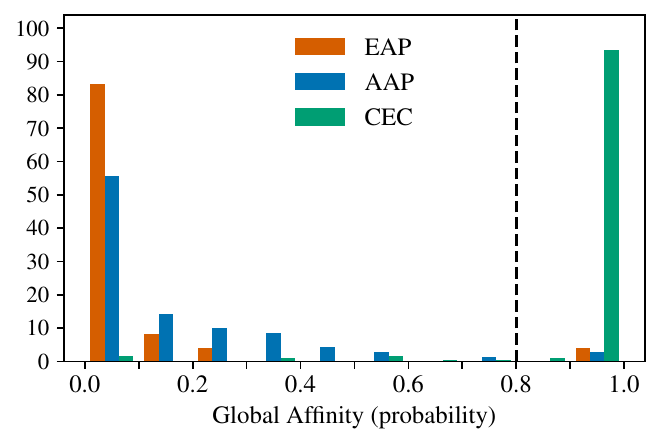} 
    \caption{
    This figure is the same as \autoref{fig:cec_hist_so}, but without correction of mislabeled examples.
    As only 5 examples differ, the plots are very similar.
    }
    \label{fig:cec_hist_so_uncorrected}
\end{figure}
Here we discuss the misclassified examples in the \citet{zhou-etal-2024-constructions} dataset.

Our approach when run on the original dataset appears to misclassify 11 examples (1 EAP, 2 AAP, 8 CEC).
As we detail here, only 5 of these are actually misclassified.
\autoref{fig:cec_hist_so_uncorrected} shows the same histogram as presented in the main text (\autoref{fig:cec_hist_so}) but without any corrected labels or omissions.

For each apparently misclassified sentence, we categorize as 
\begin{enumerate}
    \item \textbf{Misclassified}: Our (untrained) classifier was wrong; the original label was correct. No changes are made.
    \item \textbf{Mislabeled}: Our classifier identified a mislabeled example. We correct the label and present the analysis in the main text with the corrected label.
    \item \textbf{Unclear}: An example that is sufficiently ambiguous in interpretation to be omitted or linguistically invalid. 
    We omit the example from analysis in the main text.
\end{enumerate}

\subsubsection{CEC with prob < 0.8}
There are 9 CEC examples originally labeled as CEC that appear to be misclassified (low global affinity on \ttw{so}).
Of these 9, 2 are mislabeled and 1 is linguistically invalid.\\

\noindent
\textbf{Mislabeled or Unclear:}

\begin{enumerate}
\item
\ttw{It's his lucky quarter and Pop feels so bad that Lucky lost it.}\\

\textbf{Mislabeled} (Relabeled to AAP from CEC)

\item 
\ttw{I am so fortunate to have had it recommended to me so highly that I bought the eight pack.}\\

This sentence is \textbf{unclear} and does not seem to be an instance of any of the three constructions.
We \textbf{omit} it.

\item
\ttw{I am so ashamed of myself that I ignored other reviewers and paid money for this book.}\\

\textbf{Mislabeled} (Relabeled to AAP from CEC)
\end{enumerate}

\noindent
\textbf{Misclassified}
\begin{enumerate}
\item
\ttw{It has also been noted that he was so satisfied that he did this without fee or reward and was publicly thanked.}\\

This example is ambiguous:
It can be read as a CEC (``so satisified'' $\rightarrow$ ``did this without fee'') or as an EAP (he was satisfied that he was thanked).
We conservatively leave this example in the dataset.

\item
\ttw{The judges were so surprised that one of them had a "spasm," one leaned against the wall for support, and the other fell backwards into a barrel of flour!}\\

This is a valid example, although an affective interpretation is possible. 

\item
\ttw{But, a friend was so adamant that I tried it.}

\item
\ttw{I was so confident that I made changes on my own.}

\item
\ttw{There are a couple of false notes along the way, such as a dreadful rendition in front of a room of people of "You're So Vain," but so many moments are so right that I had no trouble forgiving them the few missteps.}

\end{enumerate}

\subsubsection{AAP with prob > 0.8}
There are 2 sentences labeled as AAP with probabilities > 0.8.
One is mislabeled, and the other is unclear.

\begin{enumerate}
\item
\ttw{This was so funny that I had to buy another copy and read it to my better half.}\\

\textbf{Mislabeled} (Relabeled to CEC from AAP)

\item
\ttw{After a series of fires in 1741, the city became so panicked that blacks planned to burn the city in conspiracy with some poor whites.}\\

This sentence is \textbf{unclear}. 
As written the sentence suggests that the panic may be \emph{causing} the plan to burn down the city (hence the CEC interpretation).
However, a closer review of the entire sentence suggests that the intended meaning is that a series of fires (potentially arson) led the city to be afraid that there was a conspiracy to commit further arson.
The best expression of intended AAP meaning could be achieved by removing \ttw{so}, and changing verb tense:

\ttw{After a series of fires in 1741, the city became panicked that blacks \ttw{were planning} to burn the city in conspiracy with some poor whites.}

We \textbf{omit} this sentence.

\end{enumerate}

\subsubsection{EAP with prob > 0.8}
The 1 EAP example with probability > 0.8 is unclear.
\begin{enumerate}

\item
\ttw{In Burma, the belief was once so widespread that the Sumatran rhino ate fire.}\\

Arguably \textbf{unclear}.
For one of the authors, this sentence seems to suggest (nonsensically) that the belief's being widespread was the \emph{cause} of the rhino's eating fire.
The \ttw{so} could be replaced with, e.g., \ttw{surprisingly} to achieve the apparently intended meaning. 
Alternatively \ttw{so} could simply be omitted.

We \textbf{omit} this sentence.

\end{enumerate}

\subsection[Supplement for Section~\ref{cec:so-that}]{Supplement for \sectionref{cec:so-that}}
\subsubsection{Dataset (multiple-that)}
\label{app:so-that}

To produce the augmented multiple-that dataset, we searched for existing examples in the dataset \cite{zhou-etal-2024-constructions} that already had two or more \ttw{that} words. 
In some cases, we insert additional complementizer (\ttw{that\ldots}) phrases.
Some examples are created with two CEC phrases to test that each \ttw{so} has high local affinity with its associated causal \ttw{that}.
We label the correct causal \ttw{that} for the analysis.

We provide two examples from the 31-sentence multi-that dataset:
\begin{enumerate}
    \item 
    This example has 5 \ttw{that}-phrases. 
    The affinity of \ttw{so} with the correct \ttw{that} is more than two orders of magnitude higher than with any other \ttw{that}.\\
    
\ttw{John worked \textbf{}{so} hard on helping his friend improve his argument that the policy was bad and that America should adopt the resolution that the policy had failed \textbf{that} he was too tired to debate the topic that the policy had failed himself}.

\item
Some examples are double CEC with multiple \ttw{so-that} pairs. 
We test that both each \ttw{so} has the highest affinity with the correct \ttw{that}. For example:\\

\ttw{Li shiji was \textbf{so$_1$} thankful \textbf{that$_1$} he wept and bit his finger \textbf{so$_2$} hard \textbf{that$_2$} he bled.}

\end{enumerate}

\subsection[Supplement for Section~\ref{cec:eap-aap}]{Supplement for \sectionref{cec:eap-aap}}
\label{app:cec-umap}

\begin{figure}[t] 
    \centering
    \includegraphics[width=\columnwidth]{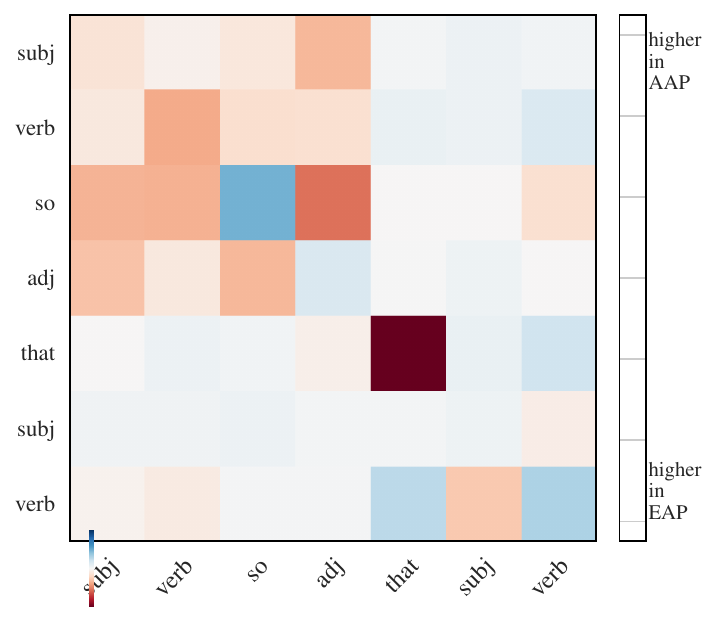} 
    \caption{This plot shows which interactions, \emph{on average}, most substantially differ between EAP and AAP examples.
    From highest to lowest in absolute value we have \ttw{that} with itself, \ttw{so} with \ttw{adj}, \ttw{so} with itself, \ttw{verb} with itself, and \ttw{so} with \ttw{verb}.
    }
    \label{fig:cec_aap_eap_diffs}
\end{figure}

\begin{figure}[t] 
    \centering
    \includegraphics[width=\columnwidth]{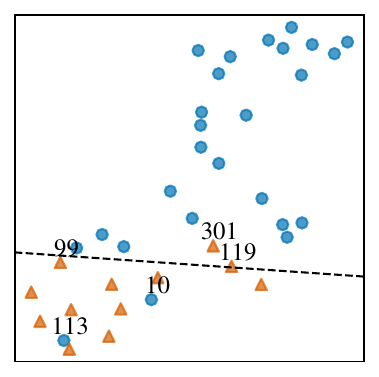} 
    \caption{This figure is the same as \autoref{fig:cec_aap_eap_umap} but with labels for examples which would likely cluster with the wrong class or are on the boundary.}
    \label{fig:app_cec_aap_eap_diffs_labels}
\end{figure}

\begin{figure*} 
    \centering
    \includegraphics[width=\textwidth]{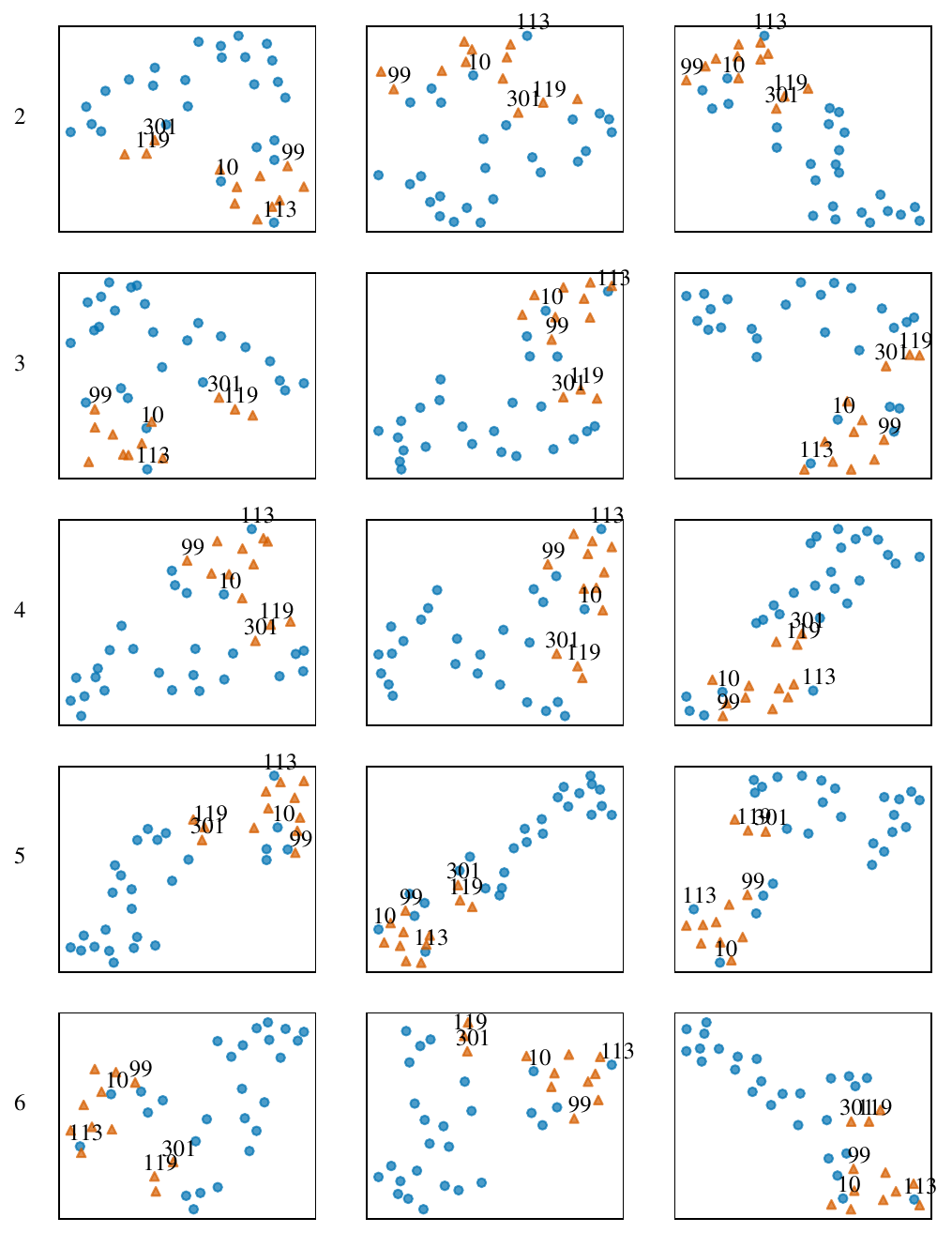} 
    \caption{
    Multiplot version of \autoref{fig:cec_aap_eap_umap}.
    UMAP projections for EAP (orange) and AAP (blue). 
    Each row corresponds to the number of dimensions that are used for the projection (2-6). 
    Columns correspond to different random seeds. 
    Potentially ``misclassified'' sentences (those that were near the class boundary) are labeled with their original dataset IDs for discussion.
    }
    \label{fig:umap_multi}
\end{figure*}

As described in \sref{cec:eap-aap}, we identify potentially salient differences in the EAP and AAP patterns: 
we compute a single average (position-wise) affinity matrix for all EAP examples and another average matrix for all AAP examples.
We fill the diagonal of the matrices with the average global affinity score (as probability).
We then subtract these two average matrices and then take the position-wise absolute value.
The largest values in the resulting matrix provide the interactions that seem to most distinguish the EAP and AAP. 
The heatmap in \autoref{fig:cec_aap_eap_diffs} shows the most potentially informative interactions across EAP and AAP examples.
Here we see that the most different interactions (between EAP and AAP examples) on average are \ttw{that} with itself, \ttw{so} with \ttw{adj}, and \ttw{so} with itself.

After running the pre-processing pipeline for tokenization we have 68 AAP and 24 EAP examples (see \ref{app:cec-data}).
For our analysis, we exclude any that have multitokenized words in one of the seven positions we consider, 
or for which Spacy \cite{spacy} fails to label a POS.
This leaves us with 14 EAP and 26 AAP examples for the analysis.

All UMAP projections use \texttt{n\_neighbors} 10 and \texttt{min\_dist} 0.1.
\autoref{fig:umap_multi} shows UMAP plots using different numbers of affinities: 
each row corresponds to the number of informative dimensions that are chosen using the heatmap.
The two columns show two random seeds for the UMAP projection.
From the multiplot, we see that using five dimensions produces seems to mostly separate the two classes, and  we use five dimensions in the main text (\autoref{fig:cec_aap_eap_umap}).

Finally, we label five sentences in the plots (\autoref{fig:umap_multi}) which either look to violate separability or which are on the boundary.

\begin{enumerate}
\item
Label \emph{10}: \ttw{It's his lucky quarter and Pop feels so bad that Lucky lost it.}\\
After our CEC correction (relabeled to AAP from CEC), this would likely be misclassified using UMAP, since it clusters with EAP.

\item
Label \emph{99}: \ttw{An hour later, however, they're still alive which confuses Elijah and Rebekah, as they were so positive that Klaus originated their bloodline and were sure it wasn't Kol Mikaelson (Nathaniel Buzolic).}\\
This is epistemic, and could be misclassified using UMAP.

\item 
Label \emph{113}, \ttw{In the police court, Mrs. Jones says she was so shocked that her husband had the box.}\\
This is affective and would likely be misclassified using UMAP.

\item
Label \emph{119}: \ttw{I am so sure that the lack of men on stage made some men feel excluded.}\\
This is epistemic, and could be misclassified using UMAP.

\item
Label \emph{301}: \ttw{I am so optimistic that I made the best choice.}\\
This is labeled as epistemic, though \ttw{optimism} conveys some degree of affect.
This label is arguably ambiguous.

\end{enumerate}

\section{Supplement for CoGS (\autoref{sec:cogs})}
\label{app:cogs}
CoGs has the following counts of examples for each construction type:
\begin{enumerate}
    \item \textbf{Causative-with:} 50 (for \ttw{with})
    \item \textbf{Comparative correlative:} 54 (for \ttw{the})
    \item \textbf{Conative:} 51 (for \ttw{at})
    \item \textbf{Let-alone:} 51  (for \ttw{let} and \ttw{alone})
    \item \textbf{Much-less:} 50  (for \ttw{much} and \ttw{less})
    \item \textbf{Way-manner:} 54 (for \ttw{way})
\end{enumerate}

We did not have any errors, so the number of examples reported in \autoref{fig:cogs-boxplot} are exactly these, except for the comparative correlative ``\ttw{the}'', for which there are $2\times54 = 108$, since we treat the two \ttw{the} as a single class in the analysis.

\section{Supplement for MAGPIE (\autoref{sec:magpie})}
\label{app:magpie}
\subsection{Data Sample}
Here we provide two examples drawn from the MAGPIE dataset for \ttw{nuts and bolts}:
\\
\textbf{Literal usage}: They would include orders for routine raw materials such as steel stock; screws; \emph{nuts and bolts}; lubricants and fuel oil.
\\
\textbf{Figurative usage}: Jay comes from a different end of the spectrum to Dave Ambrose, but the two both like to talk \ttw{nuts and bolts}.

\subsection{Methods}
\label{app:magpie:methods}
Each of the 49,395 sentences in MAGPIE has a PIE that is labeled as either figurative or literal.
The words that participate in the PIE are annotated with character spans. 

We omit 3,944 sentences for which annotation confidence (figurative or literal) is < 99\%. 
We omit 2,016 where labeled word offsets are wrong (i.e. the indicated word does not match the characters in the span).
This gives us 45,450 sentences and 117,385 individual PIE word spans (roughly 2.6 words per PIE) for the analysis. 
Of these, 10,313 sentences (23,484 spans) are literal and 34,138 (95,917) are figurative. 
Using RoBERTa, 3,556 words were multitokenized and there were 39 other errors. 
These were omitted.
Our analysis is conducted on the remaining 113,790 singly-tokenized words.

For each labeled character span that is part of a PIE, we simply calculate the global affinity and treat it as a classification signal for figurative versus literal use.
Our reported AUC of 0.71 in the main text includes only dataset entries in which the example sentence is at least ten words long (sufficient context), and where the individual word is at least four characters (avoiding, e.g., determiners and other short words).
If no filtering is performed the AUC for the ROC is 0.69 (versus 0.71).

\begin{figure}[t] 
    \centering
    \includegraphics[width=\columnwidth]{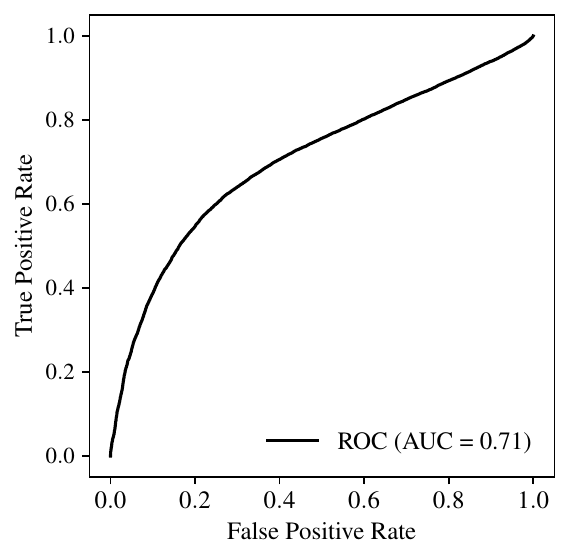} 
    \caption{ROC curve reflects informativeness of global affinity in figurative vs. literal usages of PIEs.  
        }
    \label{fig:idioms_roc}
\end{figure}

\subsection{Figurative vs. Literal Analysis}
\label{app:magpie:figlit}
\autoref{fig:idioms_fig_vs_lit2} compares per-idiom figurative versus literal averages global affinity scores.
Only idioms with at least 5 example sentences for both figurative and literal are shown (203 total). 
Idioms are sorted by figurative score.
Brackets give number of examples of each type: [\#figurative, \#literal].
Idioms where figurative (green) is higher than literal (orange) suggest ``success'' of the method (e.g., \ttw{in a rut}, \ttw{nuts and bolts}).

Though we do not conduct a full analysis of MAGPIE in this study, we report a few examples from qualitative analysis.
In the same way that affinity was able to help identify mislabeled examples in the CEC dataset, affinity draws attention to certain issues or areas of interest in MAGPIE.

For example, consider \ttw{join the club} (0.20 fig vs. 0.49 lit), which we examine to understand why the literal score is higher than the figurative.
MAGPIE's literal examples include usages of the form \ttw{join a club} (e.g., \ttw{A player joining a new club\dots}).
In figurative usages, \ttw{join the club} does not generally admit of lexical or syntactic modification.
High affinity for literals reflects contextual activation of \ttw{join a club} or \ttw{join a/an X club} type usages.
Moreover, of the few (6) figurative usages in the dataset, many are quoted discourse (\ttw{``Join the club,'' said Connie}), which do not produce sufficient activation since other completions like \ttw{``Join the group''} would be valid (hence low affinities).
In this dataset, \ttw{join a/an X club} literal usages tend to be longer and better formed. 
A fairer comparison might enforce a common context length. 

Similarly, for \ttw{play the field} (0.22 fig, 0.42 lit;  figuratively meaning to hold an interest in a number of people or things), many literal usages refer to a \ttw{playing field} (for sports). 
The syntactic patterns clearly differ, and affinity has no mechanism to attend to this. 
For example, \ttw{playing fields} in \ttw{\textbf{playing fields} and football pitches} is entrenched as its own collocation that is likely unrelated to the entrenchment of the figurative \ttw{play the field}.

Sometimes low scoring figurative usages or high scoring literal usages are mislabeled. 
For example, the top scoring literal example for \ttw{in black and white} (0.92 fig, 0.90 lit overall) is actually mislabeled as literal: \ttw{\ldots but the complicated plot is hard to follow and the characters are starkly drawn in \textbf{black and white}}.

\newlength{\halfwidth}
\setlength{\halfwidth}{\dimexpr 0.5\textwidth - \tabcolsep}
\begin{figure*}
\begin{tabular}{@{} p{\thswidth} p{\thswidth} p{\thswidth} @{}}
    \raisebox{-\height}
        {\includegraphics[width=1\linewidth]{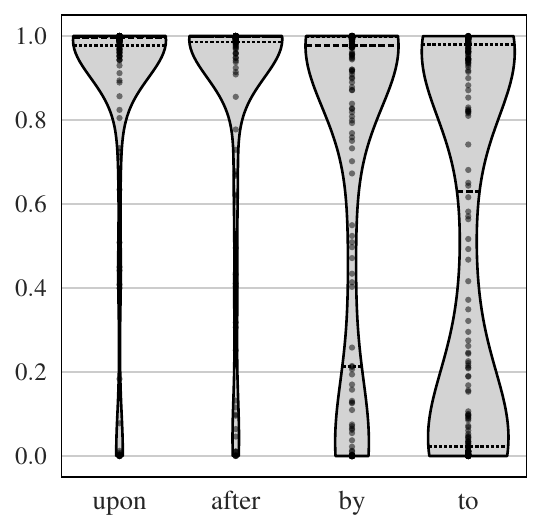}}&
    \raisebox{-\height}
        {\includegraphics[width=1\linewidth]{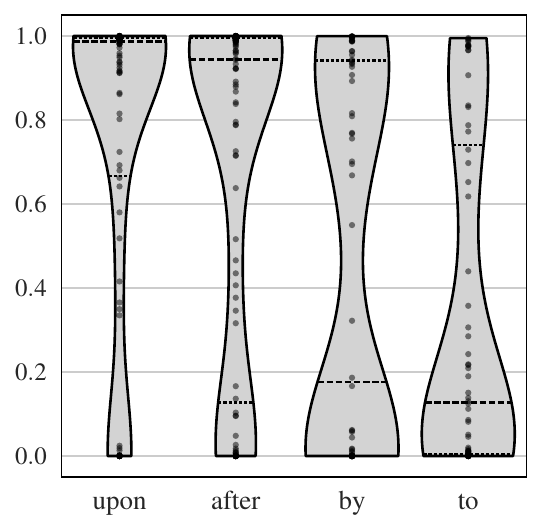}}&
    \raisebox{-\height}
        {\includegraphics[width=1\linewidth]{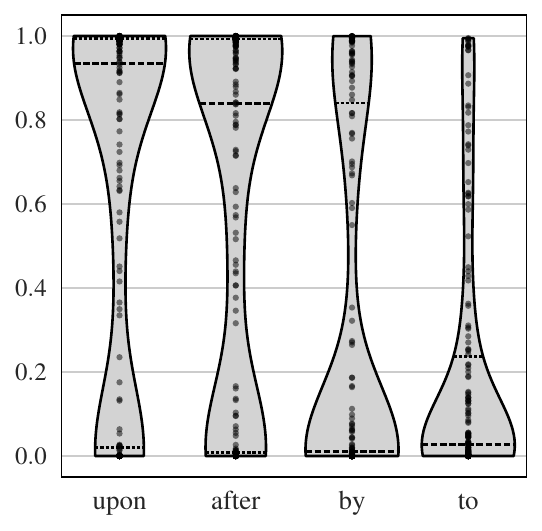}}
    \\
    \caption{
    Same as \autoref{fig:npn-violin-acceptable}, but with all sentences shown (no acceptability filter). 
    Scores skew lower, reflecting that less acceptable sentences have lower global affinities.
    }
    \label{fig:npn-violin-all} &
    \caption{
        Global affinity for NPNs using the challenge dataset of entirely unattested NPNs with acceptability $\geq4$.
        Compare to the non-challenge result shown in  \autoref{fig:npn-violin-acceptable}.
    }
    \label{fig:npns_challenge_acceptable} &
    \caption{
        Same as \autoref{fig:npns_challenge_acceptable} but with all sentences shown (no acceptability filter).
    }
    \label{fig:npns_challenge_all} 
\end{tabular}
\vspace{-30pt}
\end{figure*}

\section{Supplement for NPNs (\autoref{sec:npn})}
\label{app:npn}

\subsection{Methods}

We use GPT-4 via the OpenAI API, version gpt-4-0613, temperature 0.7, max tokens 100.
Total cost to produce 400 sentences is less than \$5.
We prompt as follows, where ``\{phrase\}'' is the particular targeted NPN (e.g., \ttw{day by day}):
\begin{quote}
An NPN construction is one like "day by day" or "face to face". 
It has a repeated singular noun with a preposition in the middle.
Other prepositions are also possible: "book upon book", "week over week", "year after year". 
Please use "\{phrase\}" in an NPN construction, 
placing "\{phrase\}" in the middle of the sentence. 
Make sure the sentence establishes a context in which the noun makes sense.
Please provide only the sentence in the response.
\end{quote}
We verify that generations match the desired form noun+prep+noun.

To obtain acceptability judgements, we randomly sort all sentences and the last author annotates with a score between 1 and 5, inclusive.
During acceptability judgement, 2 of the 100 sampled nouns were deemed to be inappropriate, and thus we omitted (2 $\times$ 4 = 8) of the generations. 

We produce two datasets:
The first is a random sample from all of RoBERTa's singly-tokenized, singular nouns.
The second uses the same sampling procedure but rejects any sampled noun for which the infinigram API has a non-zero count in Pile-train for any of the four NPNs (i.e., the sampled noun with any of the four prepositions).

\subsection{Infinigram and the Pile}
\label{app:npn:infinigram}
RoBERTa is trained on five corpora and comprises 160GB of text.
Rather than recreating the RoBERTa dataset for our frequency analysis, we search the Pile-train (800GB) using the infinigram API.
The Pile is roughly five times larger than RoBERTa's training data and the five datasets on which RoBERTa was trained are likely included, or partially included, in the Pile:
\begin{itemize}
    \item A newer version of BookCorpus is included in Pile
    \item English wiki is included in the Pile
    \item CC-news is likely included in the Pile-CC, and we verify this by spot-checking CC-news examples using infinigram, finding that most queries are successful
    \item A newer version of OpenWebText is included
    \item Stories, as part of CommonCrawl is likely included in the Pile-CC, though spot-checks do not find all queried strings
\end{itemize}

\subsection{NPN Dataset: Random sample}
When filtered to sentences with acceptabilities of at least 4, we have 
244 sentences (upon: 65, after: 73, by: 52, to: 54) and 488 nouns (two for each sentence).

\subsection{NPN Dataset: Zero frequency in the Pile}
We initially sample 94 nouns and censor 3 of them giving 91 nouns.
We generate 364 sentences.
When filtered to sentences with acceptability of at least 4, we have 
171 sentences (upon: 51, after: 54, by: 34, to: 32).

\autoref{fig:npns_challenge_acceptable} shows that affinities are lower for this challenge set, but most NPNs using \ttw{upon} still have high affinities, and over half of NPNs using \ttw{after} have high affinity (median affinity $\geq 0.9$.
This provides good evidence for generalization of the NPN with these prepositions.
NPNs using \ttw{by} and \ttw{to} have lower affinity scores, which accords with the view that they are less productive. 
Again we observe lower overall affinities when we include sentences judged to be unacceptable (\autoref{fig:npns_challenge_all}).

\subsection{Dataset examples}
We provide selected examples from our dataset:
\begin{enumerate}
    \item 
    
Generation in the random sample with high acceptability:
\\
\ttw{As a diligent scholar, he poured over his research, analyzing \textbf{manuscript} after \textbf{manuscript} to ensure the accuracy of his findings.}

Acceptability 5; affinity scores: 99.7, 99.7.

\item
Generation in random sample, with low acceptability:
\\
\ttw{Through the philosophical discussions, they delved deeper into the subject, unraveling \textbf{ambiguity} by \textbf{ambiguity}, until clarity was achieved.}

Acceptability 1; affinity scores 99.1, 96.2.

\item
Generation from the challenge dataset with high acceptability:
\\
\ttw{They lived a nomadic life, moving from \textbf{resettlement} to \textbf{resettlement}, always searching for a place to call home.}

Acceptability 5; affinity scores 98.4, 96.0.

\item
Generation from the challenge dataset with low acceptability:
\\
\ttw{The two rival politicians went \textbf{ire} to \textbf{ire} in a heated debate.}

Acceptability 1; affinity scores 0.0, 0.0.

\end{enumerate}

\section{Supplement for Comparative Correlative (\autoref{sec:cc})}
\label{app:cc}
For each sampled word, we substitute it into the original sentence and use Spacy to check whether it is a comparative adverb or comparative adjective.
To calculate the percentage of the output distribution nucleus that is a comparative adj/adv, we order the outputs by probability and iterate through them until reaching a total probability mass of $p\geq0.98$ (a nucleus using 0.98).
The final score is the proportion of the sample (the 98\% nucleus) that is a comparative adjective or adverb.

Of the 108 ($=54 \times 2$) candidate slots, 99 of them are singly tokenized and thus amenable to study using our methods.

\section[Supplement for Section~\ref{sec:characterizing}]{Supplement for \sectionref{sec:characterizing}}
\label{app: qualitative}

\begin{figure}[t] 
    \centering
    \includegraphics[width=\columnwidth]{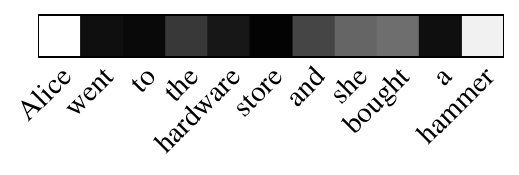} 
    \caption{
    Corresponding \emph{global} affinity plot for the local affinity plot in \autoref{fig:affinities}. 
    \ttw{hardware} has global affinity (probability) of 0.92, and is mostly affected by \ttw{hammer}, a \emph{contextual} rather than \emph{constructional} constraint. 
    See discussion in \sref{sec:characterizing}.
    }
    \label{fig:gaf_alice_hammer}
\end{figure}
\begin{figure}[t] 
    \centering
    \includegraphics[width=\columnwidth]{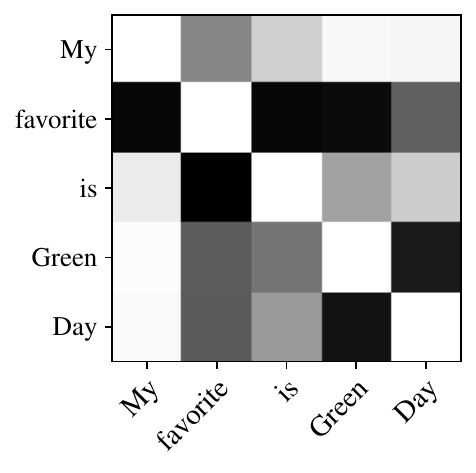} 
    \caption{Affinity matrix for Green Day without \ttw{band}. 
    Compare \autoref{fig:cx_green_band_dist}.
    }
    \label{fig:cx_green_dist}
\end{figure}

\begin{figure}[t] 
    \centering
    \includegraphics[width=\linewidth]{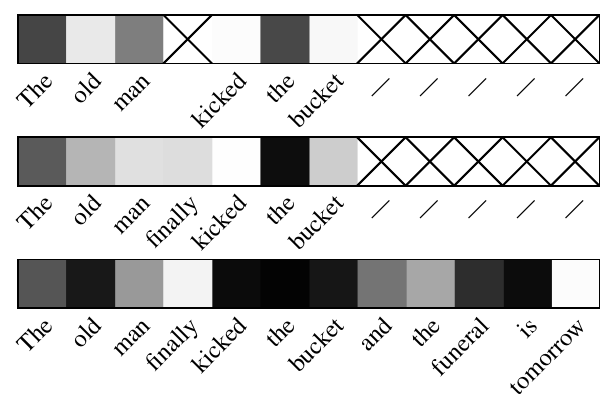} 
    \caption{
    Idioms are not always activated by context.
    \ttw{kicked the bucket} does not have high global affinity until it is clear that the old man has died. (The squares for \ttw{kicked} and \ttw{bucket} are dark only in the final sentence.)
    }
    \label{fig:kick-the-bucket}
\end{figure}

This supplement provides three additional figures.

\autoref{fig:gaf_alice_hammer} provides the global affinity plot for the example in \autoref{fig:affinities}, \autoref{sec:characterizing}.

\autoref{fig:cx_green_dist} shows the affinity matrix for the \ttw{Green Day} examples when no musical context is provided (compare \autoref{fig:cx_green_band_dist}).

\autoref{fig:kick-the-bucket} provides an additional example using the idiom \ttw{kick the bucket} (compare \autoref{fig:green-day}), illustrating 
again that even substantive constructions may exhibit low global affinities when the surrounding context is insufficient to trigger them.

\section{Use of AI Assistant}
ChatGPT4o was used to produce initial versions of python matplot generation code in some cases.
Any code produced was subsequently adapted/ modified.
ChatGPT4o was not used to write any part of this paper.
Some candidate related work was found using ChatGPT4o as a search tool.

\begin{figure*} 
    \centering
    \includegraphics[width=\textwidth]{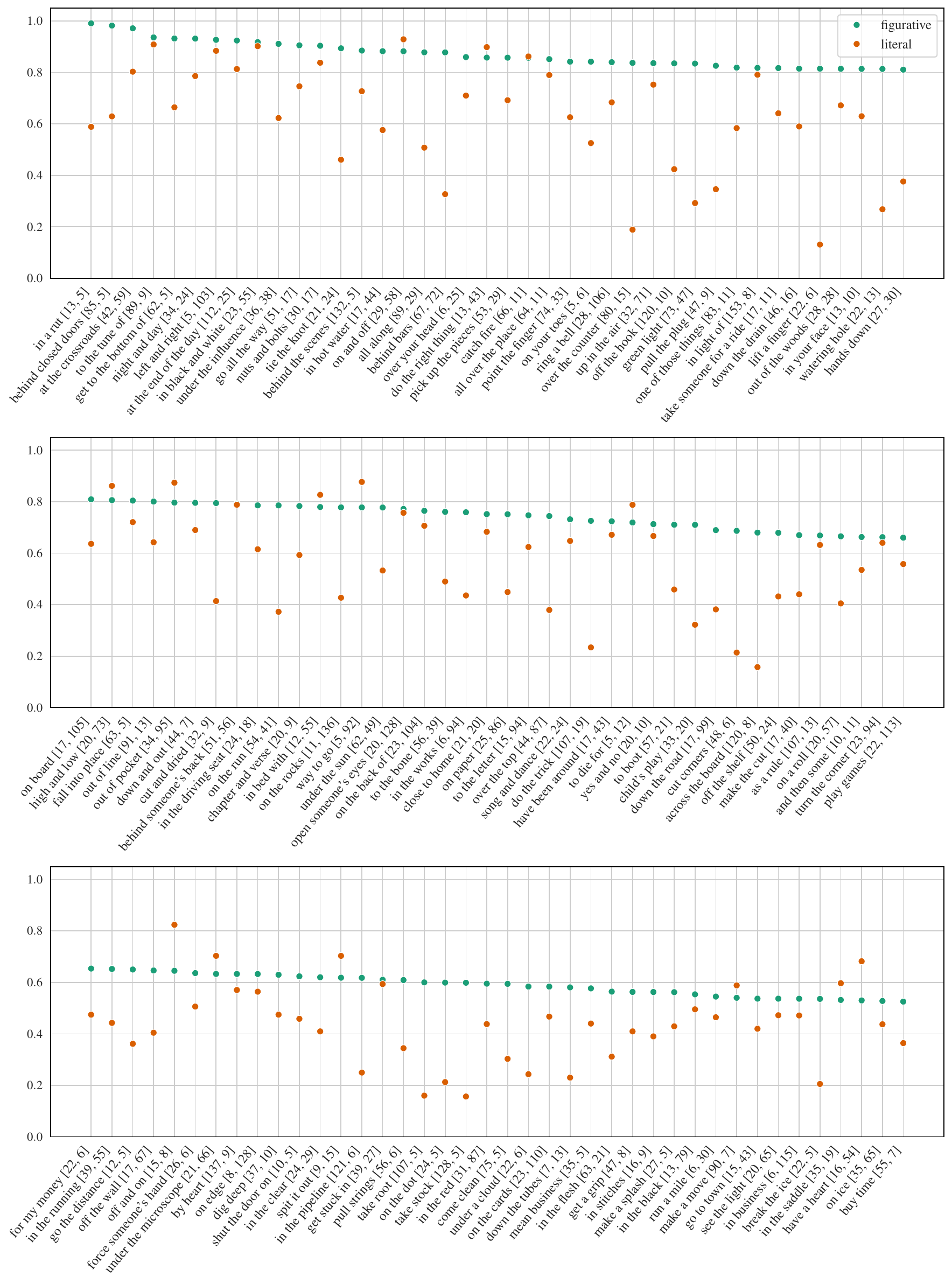} 
    \label{fig:idioms_fig_vs_lit}
\end{figure*}
\begin{figure*} 
    \centering
    \includegraphics[width=\textwidth]{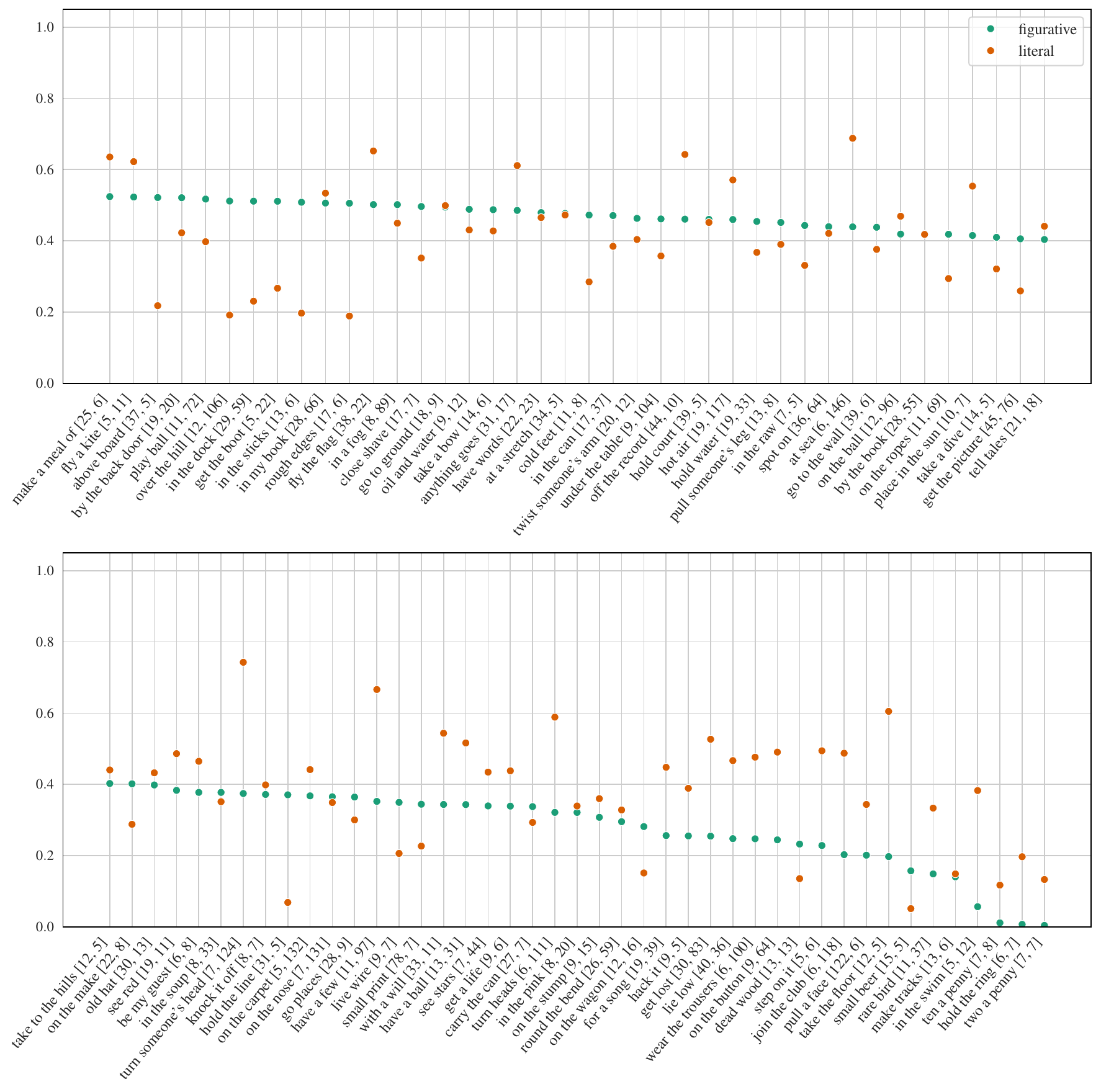} 
    \caption{
    Per-idiom global affinity scores for MAGPIE:
    average affinity for figurative uses (green) and literal uses (orange).
    Only idioms with at least 5 example sentences for both figurative and literal are shown (203 total). 
    Idioms are sorted by figurative score.
    Brackets give number of examples of each type: [\#figurative, \#literal].
    Idioms where figurative (green) is higher than literal (orange) suggest ``success'' of the method.
    }
    \label{fig:idioms_fig_vs_lit2}
\end{figure*}

\end{document}